# Gesture based Arabic Sign Language Recognition for Impaired People based on Convolution Neural Network

Rady El Rwelli[1]
Department of Arabic Language, College of Science and Arts in Qurayyat, Jouf University, Gurayat, Saudi Arabia

Osama R. Shahin[2], Ahmed I. Taloba[3]
Department of Computer Science, College of Science and Arts in Qurayyat, Jouf University, Gurayat, Saudi Arabia

***Abstract*—The Arabic Sign Language has endorsed outstanding research achievements for identifying gestures and hand signs using the deep learning methodology. The term "forms of communication" refers to the actions used by hearing-impaired people to communicate. These actions are difficult for ordinary people to comprehend. The recognition of Arabic Sign Language (ArSL) has become a difficult study subject due to variations in Arabic Sign Language (ArSL) from one territory to another and then within states. The Convolution Neural Network has been encapsulated in the proposed system which is based on the machine learning technique. For the recognition of the Arabic Sign Language, the wearable sensor is utilized. This approach has been used a different system that could suit all Arabic gestures. This could be used by the impaired people of the local Arabic community. The research method has been used with reasonable and moderate accuracy. A deep Convolutional network is initially developed for feature extraction from the data gathered by the sensing devices. These sensors can reliably recognize the Arabic sign language's 30 hand sign letters. The hand movements in the dataset were captured using DG5-V hand gloves with wearable sensors. For categorization purposes, the CNN technique is used. The suggested system takes Arabic sign language hand gestures as input and outputs vocalized speech as output. The results were recognized by 90% of the people.***

## I. Introduction

Around 60 million people use body language around the world, and an automated tool for interpreting it might have a big effect on communication between those who use it and those who don't. Sign language is a means of wordless communication that includes the use of body parts. In sign speaking and listening, face features, as well as eye, hand, and lip gestures, are used to transmit information. People who are deaf or hard of hearing rely heavily on sign language as a form of communication in their daily lives [1]. Nevertheless, the lack of consistency in shape, size, and posture of the hands or fingers in an image made computer interpretation of hand signals exceedingly difficult. SLR can be approached in two aspects: image-based and sensor-based. The main advantage of expression frameworks is that users do not need to use complicated gadgets. In any scenario, extensive operations are required during the pre-processing stage. The importance of language in growth cannot be overstated. It facilitates the internalization of social norms and the development of communication control in addition to serving as a channel for interpersonal communication. Even though they can hear the language spoken across them, deaf children do not learn a language to express themselves in the same way that hearing impairment children do.

SLR research has recently been divided into two categories: vision and contact-based approaches. This between sensing users and devices is part of the interaction technique. It usually employs an interferometric glove that collects finger motion, bending, movement, and angle information of the produced sign via EMG signals, inertial estimate, or electromagnetism. As the platform's input, the vision-based technique uses data obtained from video streams or photos captured with the camera. It's also divided into two categories: presence and 3D model-based techniques [2]. The majority of 3D model-based strategies begin to gather the position of the hand and joint angle of the hand in 3D spatial into a 2D image. Whereas demeanor identification relies on features extracted from the image's PowerPoint display, recognition is completed by matching the characteristics [3]. Although many hearing-impaired people have mastered sign language, few "regular" people understand or can use it. This has an impact on impaired people's communication and creates a sense of separation among them and the "regular" society. This chasm can be bridged by deploying a technology that constantly converts sign language to textual and vice versa. Numerous paradigm advancements in many scientific and technological fields have now aided academics in proposing and implementing systems that recognize sign languages.

Disabilities people interact through hand signals, which is a gesture-based communication strategy rather than written or spoken language. Arabic is the official language of 25 different countries. In certain nations, Arabic is spoken by only a small percentage of the population [4]. According to some accounts, the total number of countries is between 22 and 26. Although the Arabic language is deontological, the Arabic gesture is not. Arabic is spoken by Jordanians, Libyans, Moroccans, Egyptians, Palestinians, and Iraqis, to name a few. Each country, though, has its unique dialect. To put it another way, there seem to be two forms of Arabic: standard and colloquial. As they all employ the same alphabets, the Arabic sign language (ArSL) is also the same. This function is quite beneficial to research studies. The Arab deaf communities are a close-knit group. Interaction between the deaf and hearing communities is low, focusing primarily on communities with deaf people, relations of the deaf, and occasionally play companions and professionals.

The recognition of Arabic Sign Language includes a continuous identification program based on the K-nearest neighbor classifier and a feature extraction method for the Arabic sign language. However, the fundamental flaw with Tubaiz's method would be that patients must wear interferometric hand gloves to gather information on specific gestures, which can be extremely distressing again for users [5]. For the construction of an Arabic sign language recognition, an interferometric glove was created. Using hidden Markov model (HMM) and temporal characteristics, continual identification of Arabic sign language is possible [6]. A study was performed on the translation of Arabic sign language to text for usage on portable devices. While the above papers cover a wide range of sign languages, Arabic Sign Language was also the subject of research in a few instances. Using a Hidden Markov Model (HMM) quantifier, the researchers achieve 93% accuracy for a sample of 300 words. In comparison to HMM, they use KNN and Bayesian classifications [7], which produce equivalent results. This presents a network matching method for continual detection of Arabic Sign Language sentences. Decision trees and the breakdown of motions into stationary poses are used in the model. Using a polynomial runtime technique, they attain at least 63% accuracy when interpreting multi-word phrases.

This paper deals with the Gesture Based Arabic Sign Language Recognition for Impaired People and this uses the Convolution Neural Network process as the research system. There is different section that deals with the process of Arab Sign Language and the Convolution Neural Network. Section 1 organizes the Introduction of the gesture sign Arab language and the Machine Learning system. Different methods and research involved in Arab language recognition were expressed in detail in Section 2; the proposed methodology is presented in Section 3. The result and Discussion are investigated in Section 4 and finally, the paper gets concluded.

## II. Literature Review

The most comfortable and creative way for the hard of hearing to communicate is through hand signals. With improvements in multimedia tools and networks, academics have long been drawn to innovation. Sign language communications systems as a way to increase network technology for the hearing and speech impaired, offering increased social possibilities and integration. This study introduces a framework for leveraging the Microsoft Kinect device to communicate in Arabic sign language. The proposed method is based on the gesture recognition architecture for Arabic signs proposed by [8] for language communication systems. The suggested Language for Arab sign technique has a sign identification rate of 96 %, according to experimental data. In addition, the typical mission completion time for an Arabic sign was roughly 2.2 seconds. As a result, the suggested technology can be used to develop a real-time Arabic sign languages communications network. Finally, survey respondents stated that the projected procedure is consumer and simple to use and that it may be utilized to recognize and show Arabic signs at a minimal cost.

Researchers attempt to be using ICT to improve the Deaf community at large life quality by building solutions that can help them improve communication with the rest of the world and among themselves. Designers describe work on the construction of an Avatar-based translation for Deaf individuals from Arabic Speech to Arabic Sign Language in this paper. The study begins with an overview of the Deaf community's situation in the Arabic-speaking population, as well as a brief assessment related to particular research. [9] a translation system based on the avatar of Arabic speech and Arabic sign language for deaf people is recognized. The study begins with an overview of the Deaf community's situation in the Arabic-speaking population, as well as a short assessment of various related research. The next section describes the research system's design considerations. The technology will be built around a dataset of captured 3D Arabic sign language movements. Data gloves will be used to capture the gesture recognition motion.

The use of an automatic speech recognition method for Arab sign language (ArSL) has significant societal and humanitarian implications. With the growing deaf culture, such technology will aid in integrating such individuals and allowing them to live a regular life. Arab sign language, like other languages, has many subtleties and distinct qualities that necessitate the use of a useful weapon to treat it. The author in [10] propose a novel system based on deep learning that will automatically recognize words and numbers in Arab sign language when fed with a genuine dataset. Research conducted comparison research to demonstrate the effectiveness and robustness of suggested method vs established approaches based on k-nearest neighbors (KNN) and Support Vector Machines (SVM). Hearing is essential for normal language and speech development, and hearing impairment occurs whenever the acuity to normally heard noises is reduced [11]. Many studies show that one out of once each three to four educated citizens with any degree of hearing loss faces educational, social, and impede learning. The goal of this study was to assess the hearing impairment in hearing-impaired child's psychological traits (communication, social, emotional, and cognitive), and then connect this pattern to a linguistic scale.

The author in [12] proposed the Arab sign Language recognition and for those with hearing impairments, sign language entails the movement of the arms and hands as a channel of understanding. The identification of certain characteristics and the classification of specific input data are the two key steps in an automatic sign identification system. Many methods for categorizing and identifying sign languages have been proposed in the past to improve reliability. However, recent advances in the field of machine learning have prompted us to pursue more research into the identification of hand signs and gestures using deep neural networks. The Arabic gesture has seen significant research on hand gestures and gesture recognition. This research proposes a vision-based system that uses CNN to recognize Arabic hand sign-based letters and translate them into Arabic speech. With a deep learning model, the suggested system automatically recognizes hand sign symbols and shouts out another output in Arabic. This system recognizes Arabic hand sign-based letters with 90% accuracy, indicating that it is a very reliable technology. Using more powerful hand gesture recognition technologies like Motion Sensors or Xbox Kinect can enhance accuracy even more. The result will be given to the text into the voice engine, which will create the sounds of the Arabic hand sign-based characters.

Hearing-impaired people can be found throughout the world, so developing good local level sign language recognition (SLR) systems is critical. This did a thorough assessment of computerized sign language identification based on supervised learning strategies and processes published between 2014 and 2021 and found that existing systems require theoretical categorization to accurately interpret all available data. As a result, focus on aspects that are included in practically all basic sign detection methods. The author in [13] present a comprehensive framework for investigators that analyzes their advantages and weaknesses. This study also demonstrates the importance of types of sensors in this sector; it appears that acknowledgment based on the integration of datasets, such as vision-based and webcam channels, outclasses unimodal analysis. Furthermore, recent advances in research facilities have enabled them to advance from the official press of sign language protagonists and turns of phrases to the capacity to change ongoing gesture conversations with minimal latency. Many of the models available are adequate for a range of tasks, but neither of them currently has the requisite generalization potential.

## III. Methodology

### A. Arabic Sign Language Architecture

The Architecture for Arabic Sign Language Communication System is a proposed solution that serves as a low-cost multiple languages translation. A sign language-based communications network, all while retaining high precision and economical usability [7]. The following Fig. 1 Architecture of Arabic Sign Language is shown. Hardware, software, and the network are the three elements that make up the system architecture. A gesture authentication method and a video representation are included in the hardware device [14]. The Gesture recognition digital storage repository, the Sign recognition center, and the Sign media center are all part of a software component. The network device is responsible for sending and receiving Arabic sign language data across a network connection.

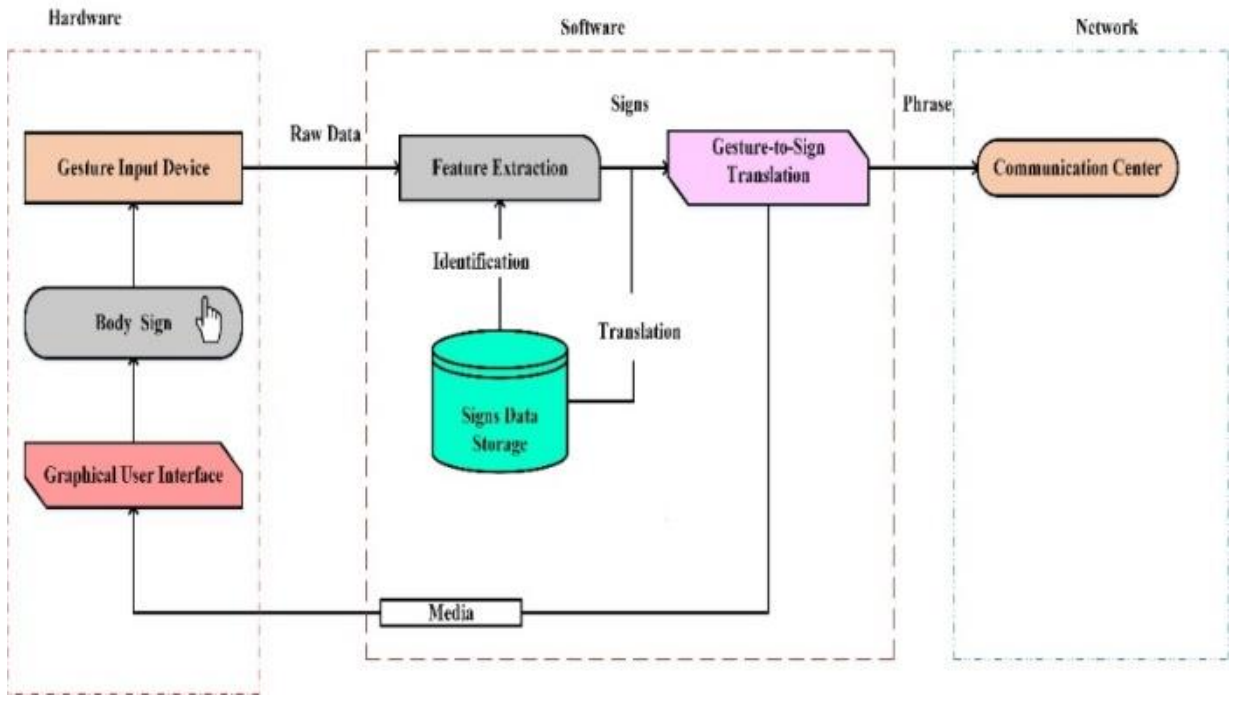

Fig. 1. Architecture of Arabic Sign Language.

*1) Hardware:* The transmitter and receiver channels via which the user controls the system are provided by the hardware components of the system. The Gesture input unit is a Microsoft Kinect from the first generations, which collects data and delivers it to the Sign recognition center [15]. The output data from the Sign media center is shown via the Display devices (audio system and digital display). In its current state, the stability supports visual data.

*2) Software*: The program's software is in charge of extracting features and movement translation, as well as offering a simple Graphical User Interface (GUI). Based on the existing and established lexicon, the Sign Identification Center turns the raw input into a collection of predefined signs [16]. The Sign language datastore is being used to obtain information about signs. The Sign media center translates the signs obtained from the Sign recognition center into the desired medium and speech, which it then sends to the Display technologies or engaging in dialogue.

The gesture recognition data store includes both gesture dictionaries and translating dictionaries from one vernacular to another. The size of something like the information storage is limited due to the project's concept stage.

Source: ArSLAT: Arab Sign Language Alphabets Translator [17].

*3) Network*: The Signs media player sends media to the Communication center, which then transmits this through the system to its intended destination [18]. This section is presently unimplemented because it is unrelated to usability testing. The above Fig. 2 shows the Arab Sign Language Alphabets and this image could be used for the further processing system.

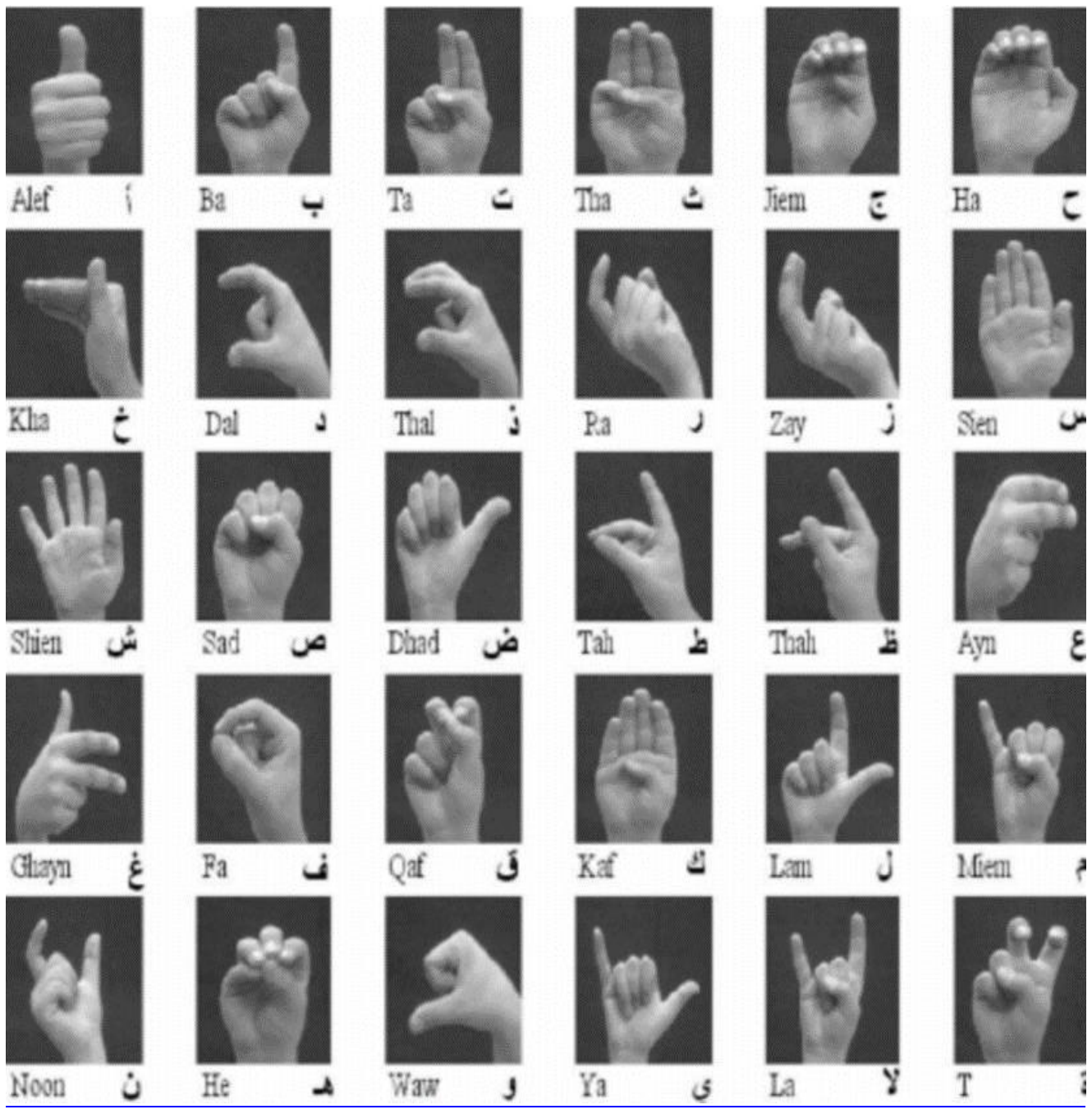

Fig. 2. Arab Hand Sign.

### B. Preprocessing of Data

The first stage in creating a working deep learning model is data preprocessing. This is used to convert raw data into a format that is both usable and efficient. The flowchart of data preprocessing is shown in Fig. 3.

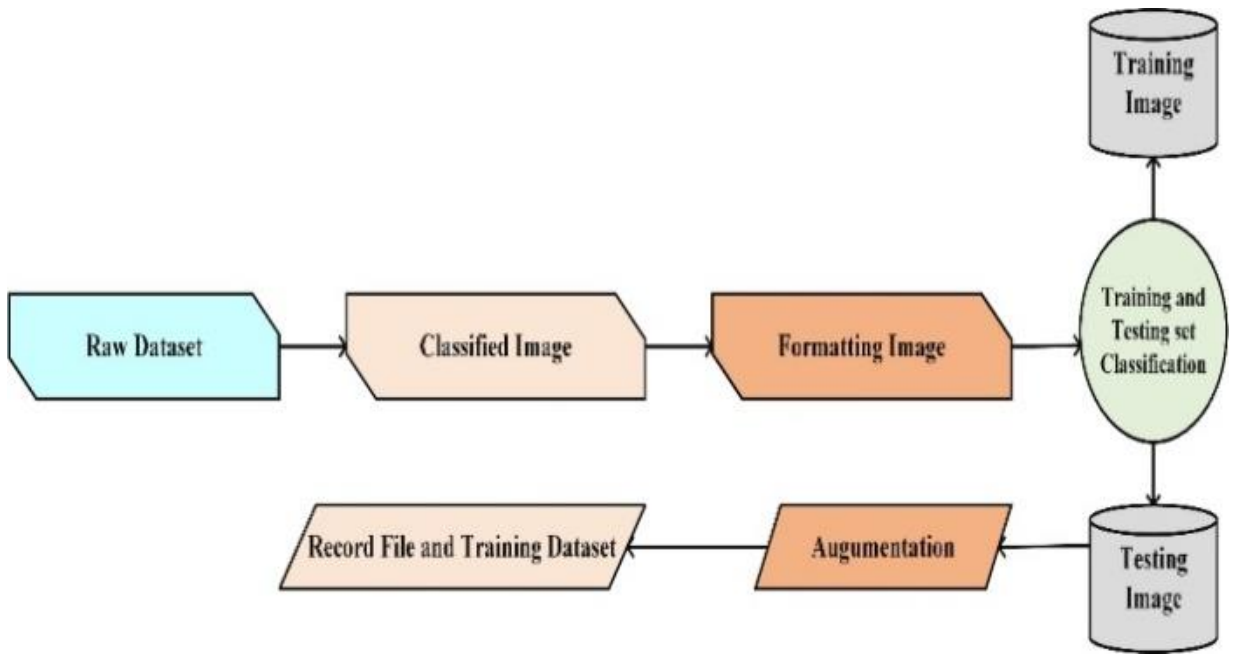

Fig. 3. Data Preprocessing.

*1) Raw data*: The image captured using the camera is termed the raw data in this section the raw image suits the hand sign image of the Arabic language and this is implemented in the proposed method [19]. The following environment is considered for image representation:

- Different angles.
- Lighting condition changing.
- Focus and good quality.
- Object size and distance adjustment.

The purpose of making raw photographs is to create a dataset that can be used for training and testing. the Arabic Alphabet from the dataset of the suggested program.

*2) Classified image*: The presented method categorizes the Arabic Alphabet's pictures. To understand the system, one subfolder is utilized to store photographs from one classification. In the developed framework, all subfolders that describe categories are maintained together in one primary special folder "dataset."

*3) Number of epochs*: The number of epochs represents how many times the complete dataset is processed into the neural network during training. There is no perfect number for it though, and it is determined by the facts.

*4) Formatting image*: The graphics of hand signs are usually uneven and have a varied background. To obtain the hand component, it is important to remove the unneeded elements from the photos. Images are referred to as digital data that has been rasterized for usage on a display device or printing in some of those types [20]. The extract was subjected is the process of converting visual data into a set of pixels.

*5) Classification of training and testing dataset*: The image taken for the formatting could be classified based on the training or testing image. A controlled learning method for classification examines the training data set to find, or learn, the best relationships between two variables that will produce a strong forecasting model. The goal is to create a trained (fitted) model that does a good job of generalizing to new, unknown data.

*6) Augmentation*: Real-time data is always incomplete and unusual due to several modifications (rotating, moving, and so on). Image augmentation is a technique used to improve the achievement of deep neural networks. It purposefully tries to manipulate images with methods such as shear, shifts, flips, and rotation. Using this image enhance raw images [21], the suggested system's images are rotated dynamically from 0 to 360 degrees. A small number of photographs were also ripped at random with a 0.2-degree range, and a small number of images were inverted horizontally.

### C. Frame Work

The design of the Arabic sign language recognition utilizing CNN is shown in Fig. 4, Convolution Layer. CNN is a machine learning (ML) system that uses perceptron algorithms in the implementation of its operations for data gathering. These systems are categorized as artificial neural networks (ANN). The discipline of machine learning is where CNN is most useful. It mostly aids in the recognition and classification of images [22]. CNN is made up of two parts: feature extraction and classification. Each element has its own set of features that must be investigated. These elements will be explained in detail in the following sections. A convolutional neural network (CNN, or ConvNet) is a type of neural network that is used to analyze visual information [23]. One of the main reasons that researchers have realized the efficacy of deep learning is the vitality of convolution layers nets in image processing. They are in charge of significant advancements in computer vision (CV), which has significant application in self-driving cars, mechatronics, unmanned aerial vehicles, safety, medical advances, and treatment options for the visual impairments.

Convolutional neural networks employ an architecture that lends itself especially well to image classification. These systems allow neural nets to learn quickly. This enables us to approach enhanced deep multi-layer systems for image classification. CNN continues to learn from data using the Backpropagation algorithm and its derived products. Modern implementations make use of specialized GPUs to improve results even further.

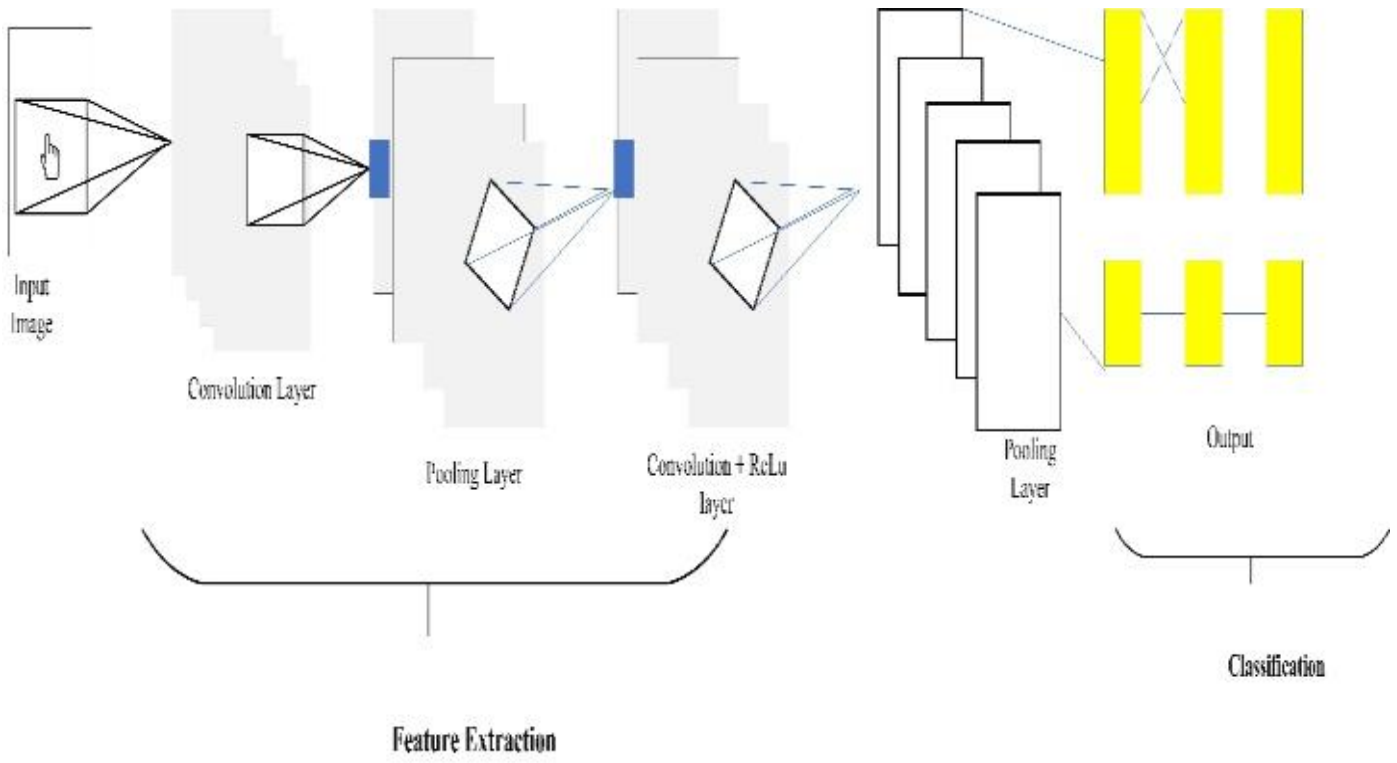

Fig. 4. Convolution Layer.

*1) Input blocks*: Squeeze Net requires an input block that would be at least 224 X 224 (With 3 channels for RGB). CNN is made up of numerous components. The Convolution operation, on the other hand, is CNN's most important component. The statistical combining of two roles to create a third function is referred to as a convolution layer. To create a feature map, the combination of the inputs using filtering or kernel is necessary. Convolution is performed by dragging each filtering over a specific input [24]. A matrix combination is performed at each location, and the outcome is added to a different feature map. Every image is turned into a 3D matrix with a defined width, height, and depth. Because the image (RGB) has color channels, the thickness is shown as a measurement.

Various convolutions can be conducted on raw data with various filters [25], resulting in various feature extraction. The output of the convolution layer is created by combining the multiple feature images. The output is then passed via an input layer, resulting in complex output. The length of a given step that the Fourier filter performs each time is referred to as the stride. The size of a step is typically 1; this indicates that the convolution filter is moving image pixel. When increasing the size of a step, the filters will slide across the input with a larger frequency, resulting in less overlap between the cells. Researchers should do something to stop extracted features from decreasing because it is always less than the input size. These are going to apply to cushion here.

$$Output_{size} = \frac{input_{size} - Filter_{size} + 2 * padding_{size}}{Stride_{\ size}}$$

*2) Max pooling layer*: In between Convolution layers, a pooling layer is naturally added. However, its primary goal is to reduce dimensionality and reduce calculation time by using fewer parameters. It also prevents overtraining and cuts down on training time. Pooling can take numerous forms, the most frequent of which is max pooling. It employs the maximum value in all windows, resulting in a smaller feature map with the same amount of information. To estimate the size of the pooling layer's output produced, the panel sizes must be specified ahead of time; the following equations can be used.

$$Output_{size} = \frac{input_{size} - Filter_{size}}{Stride_{\ size}} + 1$$

The pooling layer provides some high accuracy in all cases, indicating that a certain component will be recognizable regardless of where it appears on the panel.

The categorization component of CNN is the second most critical component. The objects are classified up of a few tiers that are all interconnected (FC). An FC layer's neurons have a strong relationship between the two to every one of the preceding layer's activations. The FC layer aids in the mapping of representations between inputs and outputs. The layer's functions are carried out using the same concepts as a conventional Neural Network [4]. One Dimensional data, on the other hand, can only be accepted by an FC layer. The flatten function of Python is utilized to create the new method for converting three-dimensional data to just one data.

*3) Dropout regularization techniques*: Overfitting is a significant and serious issue in deep neural networks. Dropout is a method of dealing with this difficult challenge. It is accomplished by randomly discarding some neural units in the neural network with an artificially designed ratio throughout training. The degree of co-adaptation between neuronal units has been greatly reduced. Using dropout on a neural network is equivalent to extracting a thinned network from the original entire network. During the training process, a series of thinning networks are collected using dropout at a specific dropout ratio. During the testing phase, it is not possible to directly help determine by combining the forecasts from exponentially thinning models. This should employ a whole untinned network with fewer weights to forecast outcomes by implicitly aggregating the results of all those thinned systems. Dropout greatly lowers overfitting and outperforms other regularization methods. The convolutional neural network using dropout would be discussed.

*4) Activation function*: The activation function is a node placed at the end of or between Neural Networks. This has various sorts of activation functions, but this discussion will concentrate on Rectified Linear Units (ReLU). The ReLU function is the most often used objective function in neural networks. ReLU has a significant benefit over other training algorithms in that it does not stimulate all neurons at the same time. The picture for the ReLU algorithm above shows that it turns all negative input to zero but does not stimulate the neuron. Because just a few neurons are stimulated at a time, it is incredibly computationally efficient. It does not reach saturation in the positive area. In reality, the ReLU activation function converges six times quicker than the $tanh$ and sigmoid activation functions.

*5) Features extraction*: The Convolutional Neural Network is made up of several basic parts. The convolution layer is a critical component of the CNN network. This layer denotes the mathematical description of functions that result in a third function. To generate a feature map, convolution must be performed within the input using a kernel or a filter. The convolution implementation consists of sliding each filter with sufficient input. At each location, matrix multiplying is conducted, and the outcome is placed on a feature map. Each image is converted to a 3D matrix with depth, height, and width. Because the image is made up of color channels, the depth has been deemed a dimension. Multiple convolutions are performed on the input dataset using appropriate criteria, resulting in distinct feature maps. The result of the convolution layer is obtained by combining the multiple feature maps. The kernel is a two-dimensional (2D) array of elements that would be used as weights generally. As demonstrated in Fig. 5, the convolution procedure is conducted by dragging the kernel across the picture. The result of the convolution layer is a feature mapping. Every section that is subjected to the dragging and convolution processes is referred to as an interesting region (IR). The accompanying equation is used to carry out the convolutional procedure.

$$Z_{ij} = (I * K)_{ij} = \sum_m \sum_n I_{i-m,j-n} K_{mn} - 1$$

Where $K$ is the kernel and $I$ is the input image. Every layer's output in Convolution Neural Network can be stated as follows:

$$y_i^l = f(z_i^l)$$

Then, $z_i^l = \sum_{j=1}^{l-1} w_i^j x_j^{l-1} - 2$

Where $y$ represents the outputs of the layer, $z$ represents the activation function, $i$ represents a layer $l$ neuron, $w$ represents the weight, and $x$ represents the input information.

$$w = \{w_{ij}^l : l = 1,3, \ldots L-1; i = 0, 1, \ldots I; j = 0, 1, \ldots J\} - 3$$

$$x = \{x_j^l : l = 1,3, \ldots L-1;\ j = 0, 1, \ldots J\} - 4$$

The pooling layer is the CCN's second layer. The primary goal of this layer is to make the feature mapping from the convolution layer easier to understand. It emphasizes the characteristic by using the maximum, summation, or averaging operations. The fully connected layer is the third layer. This layer's primary function is to transform a two-dimensional (2D) feature map into a one-dimensional (1D) one. This style is appropriate for deciding on feature categorization based on pre-defined features.

This employed an Arabic sign languages dictionary in this research. In the training phase of hand symbol identification, employed certain motions from a lexicon as a ground truth. The lexicon is distributed in the form of graphic groups of motions. Every set of images symbolizes a different type of social setting. This chooses over 40 motions performed with one hand and over 10 motions performed with both hands. This database is used to train Convolutional Neural networks. The technology was put to the test with real-life motions performed by coworkers.

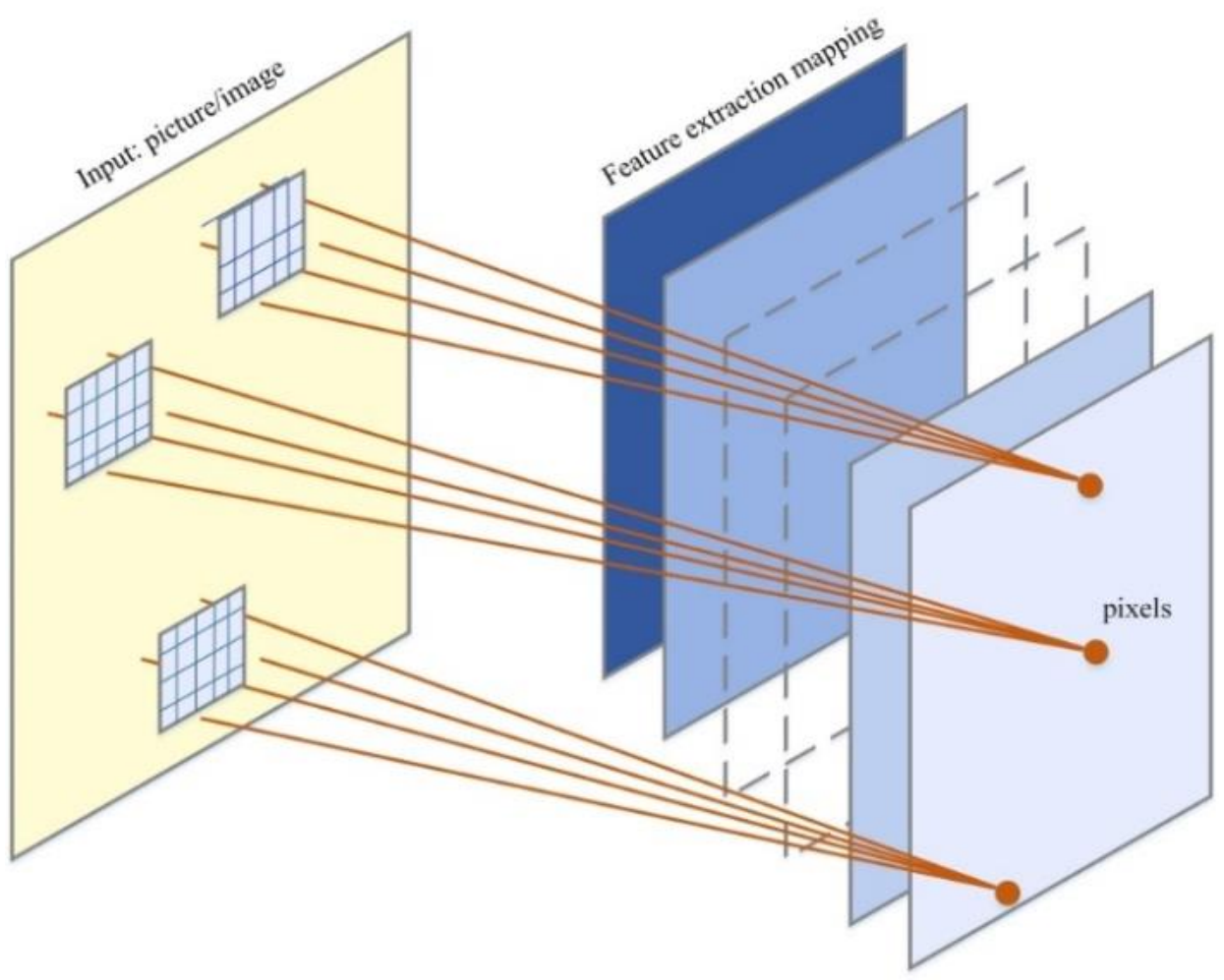

Fig. 5. Feature Extraction.

To recognize fingers and hands, computer vision architecture was created. Within its area of view, it separates and monitors them. The architecture collects movement monitoring data in the form of a series of pixels. The measured positions, sign orientations, and other information about every object recognized in the current frames are stored in the monitored data frames. A unique pixel is used to illustrate the identified fingers and hands. The flow chart for CNN proposed model is shown in Fig. 6. On still photos, summarize a technique using the methods below.

- Gesture frames are captured.
- Image denoising is a technique for improving the appearance of images.
- Employing Convolution Neural Network (CNN) to separate the face of a signer.
- Using Convolutional Neural Network to segment hand and finger gestures.
- Utilizing Convolution Neural Network to detect motions.
- By evaluating motions to elements in a pre-built database, motions can be recognized.
- Getting the translations for the motion that was recorded.

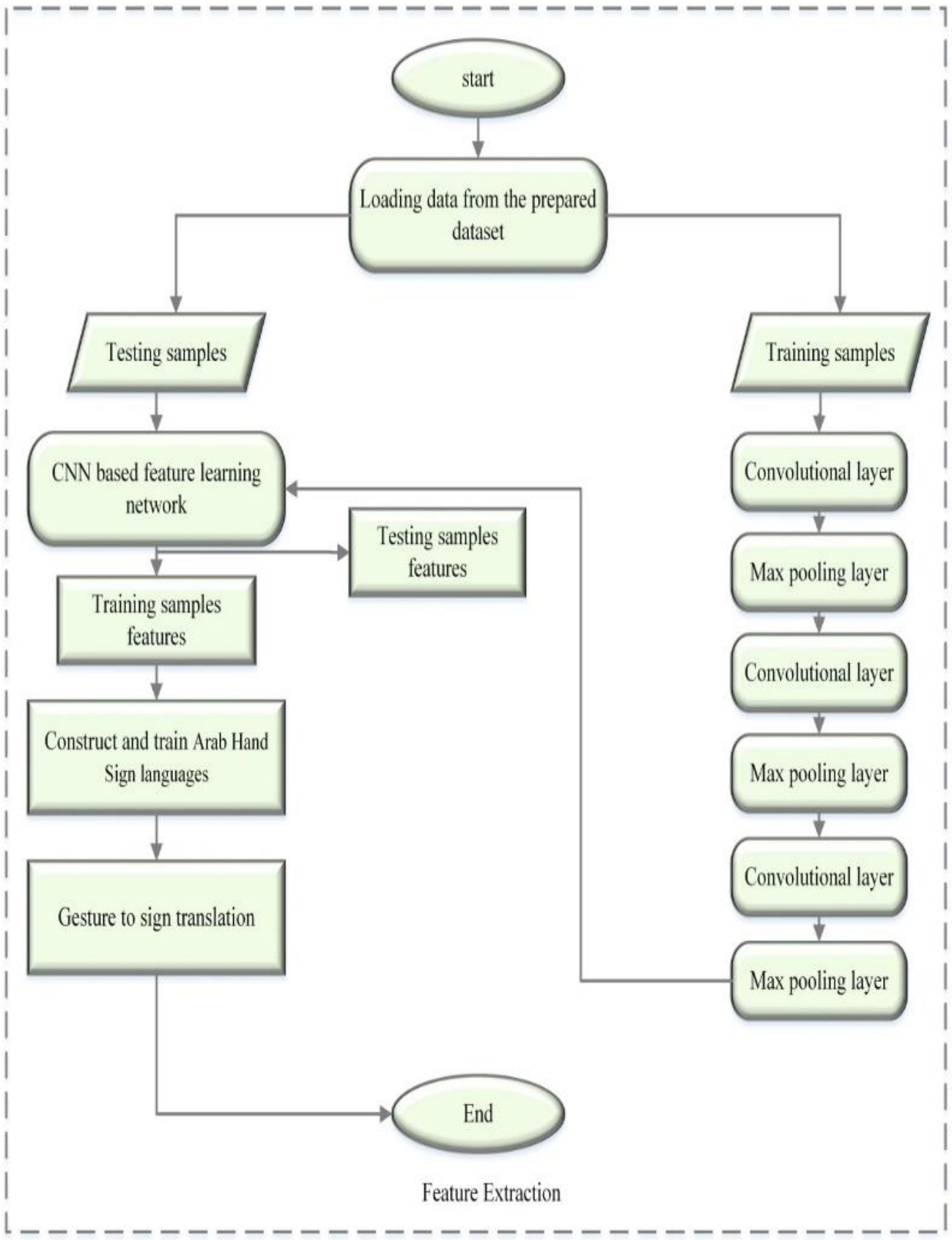

Fig. 6. Flow Chart for CNN Proposed Model.

## IV. Result and Discussion

Two convolution layers are used to test the suggested system. After that, each fully connected layer is followed by two max-pooling layers. The first layer of the convolution operation has a different pattern; there are 30 kernels in the first layer, while the second level has 64 kernels; nonetheless, the kernel size in both layers is $3 \times 3$ layer. Each pair of convolution and max-pooling was examined using two alternative dropout regularization values of 25% and 50%, respectively. As a result, this option allows for the elimination of one input out of every four inputs (25%) and two inputs out of every four inputs (50%) from each combination of convolutional and pooling layers.

The various sizes of training sets, as can be seen in Fig. 7, when training the network using 80 % of the images from the dataset, the accuracy reaches its highest point of 90.03%. Table I shows the Percentage Training Set.

Researchers compared the suggested system's outcomes with KNN (k-nearest neighbor) with Euclidean distance and SVM (support vector machines) with various kernels processors typically shown in this field to demonstrate its effectiveness.

Other factors that influence identification, including such facial movements, have been explored in prior studies. Various input detectors, like the jump action controllers, are also employed, as well as integrating multiple input detectors to handle the various characteristics described above. In addition, a novel learning approach was applied in this study, which yielded encouraging outcomes.

TABLE I. Percentage Training Set

| Training Image | Detection Rate |
|---|---|
| 60% | 85.65% |
| 50% | 83.28% |
| 80% | 90.03% |
| 70% | 88.45% |

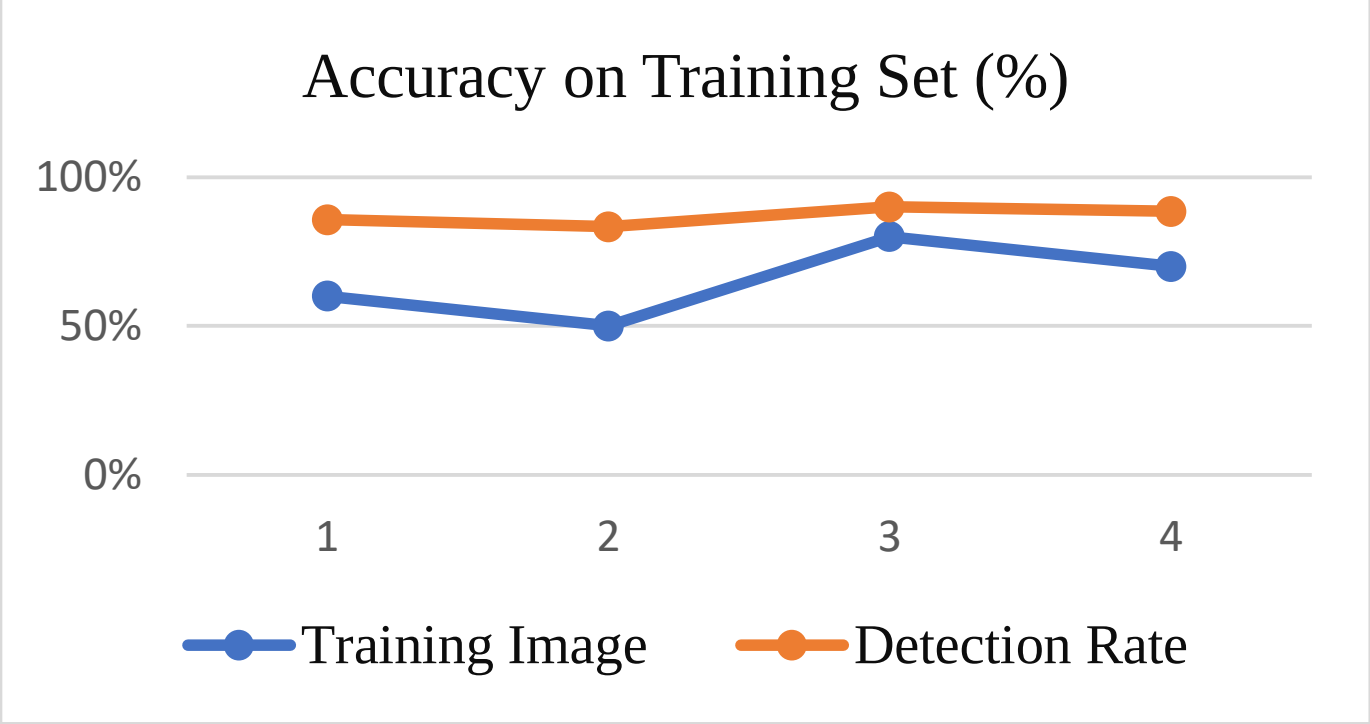

Fig. 7. Accuracy on Training Set.

TABLE II. The CNN Configuration's Categorization Results

| Input depth | CNN layer | Accuracy | |
|---|---|---|---|
| Configuration 1 | Kernel | $5 \times 5 \times 8$ | 91.5% |
| | Subsampling | $2 \times 2$ | |
| Configuration 2 | Kernel | $5 \times 5 \times 8$ | 90.6% |
| | Subsampling | $2 \times 2$ | |
| Configuration 3 | Kernel | $5 \times 5 \times 8$ | 94% |
| | Subsampling | $2 \times 2$ | |
| Configuration 4 | Kernel | $5 \times 5 \times 8$ | 92% |
| | Subsampling | $2 \times 2$ | |
| Configuration 5 | Kernel | $5 \times 5 \times 8$ | 89% |
| | Subsampling | $2 \times 2$ | |
| Configuration 6 | Kernel | $5 \times 5 \times 8$ | 89.8% |
| | Subsampling | $2 \times 2$ | |
| Configuration 7 | Kernel | $5 \times 5 \times 8$ | 86% |
| | Subsampling | $2 \times 2$ | |

The scheme then exhibits an optimistic accuracy rate with reduced loss rates in the following phase (testing phase). The accuracy rate was reduced even further when augmented graphics were used while maintaining nearly the same precision. Each digital image in the testing stage was processed before being used in this model. The proposed system generates a vector of 10 values, with 1/10 of these values being 1 and all other values being 0 to represent the predicted class value of the given data. The system is then linked with its signature step, in which a hand sign is converted to Arabic speech, Table II.

The Fig. 8 Detection Rate Comparison is shown based on Table III Comparison of Detection Rate. In this, the comparison is done between the proposed system and the other method such as SVM with RBF kernel and linear kernel and KNN. Among those methods, a higher rate of accuracy is found in the proposed method. So, by implying this method the impaired people could easily recognize the sign.

TABLE III. Comparison of Detection Rate

| Classifier | Detection Rate |
|---|---|
| Support Vector Machine with linear Kernel | 86% |
| Support Vector Machine with RBF kernel | 85% |
| KNN | 68% |
| Proposed system | 90.03% |

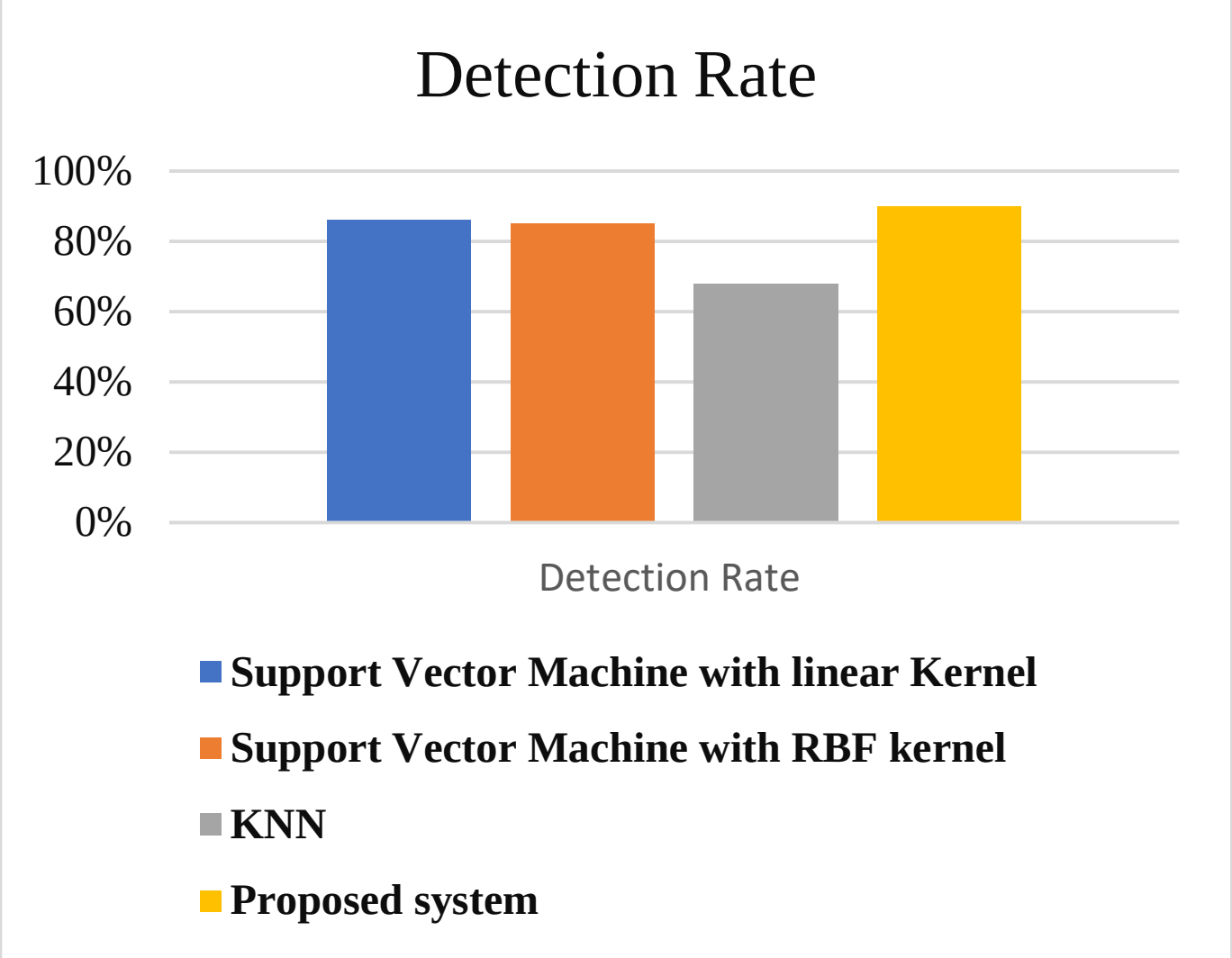

Fig. 8. Detection Rate Comparison.

## ACKNOWLEDGMENT

The authors would like to thank the Deanship of Scientific Research at Jouf University for supporting this work by Grant Code: (DSR-2021-04-0317).

Funding Statement: This work was funded by the Deanship of Scientific Research at Jouf University under grant No (DSR-2021-04-0317).

Conflicts of Interest: The authors declare that they have no conflicts of interest to report regarding the present study.

## V. CONCLUSION

The Identification of sign languages and Arabic Sign Language (ArSL), as well as several types of classifications and their outputs, were studied using various symbols and signs and ArSL. This suggested survey is carried out to take the best classifier for hand gesture recognition systems that depend on several sign languages. Some of the developed models were shown to be quite efficient, however only on limited applications. The studies originate from all across the world and include a wide range of sign language variances, which is critical for assuring worldwide coverage. The entire operation and performance of sign language recognition are represented using neural networks, machine learning, and deep learning classifiers, among others. In terms of accuracy, the Deep learning-based classifier CNN produced the results in research. Thus, the gesture Based Arabic Sign Language Recognition for Impaired People is based on Convolution Neural Network System. In addition, the size of the data collection could be enhanced further in future study projects. The suggested system's result is Arabic-language speech obtained through the detection of Arabic sign language. Furthermore, the solution presented here would be excellent for impaired people.